\documentclass[conference]{IEEEtran}
\IEEEoverridecommandlockouts
\usepackage{cite}
\usepackage{amsmath,amssymb,amsfonts}
\usepackage{algorithmic}
\usepackage{graphicx}
\usepackage{textcomp}
\usepackage{xcolor}

\graphicspath{ {images/} }

\def\BibTeX{{\rm B\kern-.05em{\sc i\kern-.025em b}\kern-.08em
    T\kern-.1667em\lower.7ex\hbox{E}\kern-.125emX}}
\begin{document}

\title{Monocular Camera Localization for Automated Vehicles Using Image Retrieval\\
\thanks{The authors are with the Model Predictive Control Laboratory,
University of California, Berkeley. E-mails: \{e.joa, fborrelli\}@berkeley.edu}
}

\author{\IEEEauthorblockN{Eunhyek Joa, Yibo Sun, and Francesco Borrelli}}

\maketitle

\begin{abstract}
We address the problem of finding the current position and heading angle of an autonomous vehicle in real-time using a single camera.
Compared to methods which require LiDARs and high definition (HD) 3D maps in real-time, the proposed approach is easily scalable and computationally efficient, at the price of lower precision.

The new method combines and adapts existing algorithms  in three different fields: image retrieval, 
mapping database, 
and particle filtering. 
The result is a simple, real-time localization method using an image retrieval method whose performance is comparable to other monocular camera localization methods which use a map built with LiDARs. 



We evaluate the proposed method using the KITTI odometry dataset and via closed-loop experiments with an indoor 1:10 autonomous vehicle.
The tests demonstrate real-time capability 
and a 10cm level accuracy. 
Also, experimental results of the closed-loop indoor tests  show the presence of a   positive feedback loop between the localization error and the control error. Such phenomena is analysed in details at the end of the article.
\end{abstract}

\section{Introduction}
Estimating the current pose of a vehicle is a key task for autonomous driving.
Since each sensor has its own limitation, multiple sensors are often utilized for localization \cite{kim2016lane}, \cite{levinson2010robust}, \cite{markoff2010google}, \cite{urmson2008autonomous}. 

Global positioning system (GPS) is a key component of localization sensor suite and one of the most common sensors used in vehicle localization applications.
While it can give reliable and accurate pose information on open roads \cite{funke2012up, zhu2006global, joa2020new}, it is well-known that GPS based localization systems fail to provide pose information on urban roads surrounded by buildings or in tunnels.

To compensate this GPS dropout issue, exteroceptive sensors such as LiDARs and cameras are additionally employed for the localization.
The common way to estimate the current pose with these sensors is a map-aided localization with LiDARs, i.e., matching the sensor measurement with pre-built high definition map \cite{chen2021range}, \cite{hata2015feature} \cite{levinson2010robust}.
While these methods show centimeter-level localization performance, they suffer from two main drawbacks. 
First, scaling high-definition map to large-scale (i.e. city-scale or country-scale) is an open challenge. 
Second, they require LiDARs for localization which are more expensive than cameras. 
Moreover, at least one camera is mounted on the majority of vehicles currently manufactured. 
Therefore, it is natural to study localization methods which can replace LiDARs with cameras for applications where high precision is not required.

An alternative way to map-aided LiDAR localization is a visual localization: estimating camera's global pose from images.
Next, we discuss two categories in visual localization: (1) 3D map based visual localization and (2) image retrieval based localization \cite{torii2019large}.

The goal of a 3D map based visual localization is to estimate the camera pose with respect to a pre-built 3D map \cite{snavely2006photo, brachmann2018learning, sattler2016efficient}. 
The map is the 3D structure often constructed using structure-from-motion method.
By using the constructed 3D structure, the camera pose can be estimated using 2D-3D matching.
While promising, this method also uses 3D map for localization.
Moreover, constructing 3D structure from images is not a trivial task and more difficult than constructing 3D structure from LiDAR point clouds. 

The goal of an image retrieval based localization (or visual place recognition) \cite{torii201524, arandjelovic2016netvlad} is to find a set of geo-tagged images from a database which are similar to the query image, and utilize the corresponding position as the position estimate. 
It is empirically shown that this approach scales to entire cities with compact image descriptors \cite{arandjelovic2016netvlad, torii201524, torii2019large}.
However, this method can only provide an approximate pose of a query image, not an exact pose.

In this paper, to overcome the current methods' limitations, we propose a new monocular camera localization method for automated vehicles.
Our proposed method requires a single camera to localize the vehicle using an offline built map database. Such database is built by using a LiDAR, a positioning system, and a camera.
The map database in our method is a database of geo-tagged images similar to Google Street View, which is easy to generate, to maintain, and to expand. 
To be specific, an entry of the map database is a compact image descriptor, extracted features, and the corresponding global pose of each image.
The localization in our method consists of three consecutive steps.
First, the localization module finds N reference images from the map database which are similar to a query image using an image retrieval method \cite{torii201524, arandjelovic2013all}. 
Subsequently, the module calculates the relative pose of the query image with respect to each retrieved reference image by solving a Perspective-n-Point problem (PnP) problem \cite{li2012robust}.
Since we have the corresponding global pose for each retrieved reference image, we obtain N hypothesis about the query image's global pose as an output of the second step.
Finally, using N hypothesis and on-board vehicle sensor signals as a measurement, we design a particle filter to estimate the global pose of the query image and thus the global pose of the vehicle.

Three are the main contributions of this paper:
\begin{itemize}
\item We introduce a new probabilistic formulation of the image retrieval method which combined with mapping database construction and real-time particle filtering allows one to  localize autonomous vehicles in real-time using a single camera.
\item We evaluate experimentally  the proposed method using the KITTI odometry dataset and via  closed-loop experiments with an indoor 1:10 autonomous vehicle.

\item We observe and study the appearance of  a positive feedback loop between the localization error and the control error when the image retrieval approach is used in closed loop.
\end{itemize}

\section{Approach}

The proposed method consists of three steps: \textit{(1) Visual vocabulary construction, (2) Mapping, (3) Localization}.

\subsection{Initialization: Feature extraction}

A starting point of all three steps is feature extraction.
There are a variety of feature detection algorithms such as Oriented FAST and rotated BRIEF (ORB) \cite{rublee2011orb}, Scale-Invariant Feature Transform (SIFT) \cite{lowe1999bobject}, and Dense RootSIFT \cite{torii201524}.
Given an image, each feature detection algorithm provides multiple feature points and the corresponding feature descriptors that describe the characteristics of feature points.

Previous works in visual localization mainly focus on increasing performance of the image retrieval and not take account of real-time operating capability.
Thus, they prefer accurate feature detectors such as RootSIFT to simple detectors such as ORB.
For example, in \cite{sattler2015hyperpoints}, the algorithm using SIFT correctly localized 91.1 \% images with precision 96.9 \% but took 3.5 hours to process 10,000 query images in offline, i.e., 1.3sec for each query image. 
The paper \cite{torii2019large} reports that the algorithm in \cite{sattler2015hyperpoints} takes 3sec for each query image.

While SIFT \cite{lowe1999bobject} or Dense RootSIFT \cite{torii201524} are used in previous works for accuracy and precision \cite{torii201524, torii2019large, sattler2015hyperpoints, brachmann2018learning}, we choose ORB feature detector for the real-time operating capability of localization.
In contrast to the previous works, our requirement for autonomous applications is  a sampling time\(\leq\)0.1sec, i.e., all localization logic must run below 0.1sec including an image retrieval step on a consumer-grade laptop with an i7 CPU and 32GB memory without any support of GPUs.
For this reason, it is desirable to choose simple, fast feature detector while sacrificing in accuracy.
While SIFT \cite{lowe1999bobject} or Dense RootSIFT \cite{torii201524} used in previous works provides a float-type descriptor, ORB provides a binary descriptor, which is much simpler and faster to extract.
In this sense, ORB feature detector is well-fitted to meet our requirement.
Moreover, in \cite{burki2019vizard} and \cite{mur2015orb} real-time capability of ORB has been  empirically proven.

The ORB feature detector extracts feature points and the corresponding feature descriptors.
Each ORB feature point denotes a position on the image plane, and each ORB feature descriptor is a 32-d vector, which entry is a 8 digit binary number.
\subsection{Step 1: Visual vocabulary construction}
Visual vocabulary is a set of representative feature descriptors that correspond to clustered ORB feature descriptors. 
The visual vocabulary construction step is conducted offline, and after the visual vocabulary is obtained, it  is utilized as an input parameter to the localization algorithm.

Visual vocabulary construction is an essential step to apply an image retrieval method.
The goal of the image retrieval is to find images from the database that are similar to the query image.
Since it is computationally expensive to compare or calculate similarities between two images, compact image representations are necessary for the image retrieval step.
By using the constructed visual vocabulary, we can compress all high-dimensional ORB feature descriptors from an image to a low dimensional matrix. 
This matrix is referred as a compact image descriptor.
The compact image descriptor used in this work is a vector of locally aggregated descriptor (VLAD) \cite{jegou2010aggregating} with intra-normalization \cite{arandjelovic2013all}, which has already shown a good retrieval performance in city-scale urban area \cite{torii201524}.

The visual vocabulary in our experiment is obtained by training the k-means as follows.
First, we collect over 1000 images to construct the visual vocabulary.
We refer a set of these images as a visual vocabulary construction dataset.
Second, we extract ORB feature descriptors from all images in the visual vocabulary construction dataset. 
Third, we train k-means with all extracted ORB feature descriptors as an input data.
k is set to 64 for k-means.
After training the  k-means classifier, all ORB features will be partitioned into 64 clusters.
The centers of 64 clusters of the k-means classifier are 64 representative feature descriptors that comprise the visual vocabulary.
In this paper, the 64 representative feature descriptors are built from over 1 million ORB feature descriptors, which are extracted from over 1000 images in the visual vocabulary construction dataset. 
\subsection{Step2: Mapping}
The map in this paper is in the form of a database, and each entry in database contains information of an image. 
The information of a image consists of the extracted ORB feature descriptors with their 3D coordinates in camera coordinate system, the compact image descriptor VLAD, and the global pose of the image. 
In this paper, we refer to all mapped images as reference images. 
To build a map for outdoor test, we use a LiDAR, a differential GPS, and a camera. 
For indoor test, we use a depth camera and an indoor positioning system instead of a LiDAR and a GPS, respectively. 
In the rest of the section, we will describe each component of the map database.

\textit{Extracted ORB feature descriptors with their 3D coordinates.} We extract a maximum of 1000 ORB feature points and the corresponding feature descriptors from each reference image. 
Since it is not possible to find relative pose between two images without known 3D coordinates of feature points \cite{szeliski2010computer}, we need an aid of sensors or algorithms to get 3D coordinates of each feature point.
To get the 3D coordinates of each feature point, we utilize LiDAR and depth camera for outdoor and indoor test, respectively.
Since the field of view of LiDAR (or depth camera) is not exactly matched with that of camera, not all ORB feature points have the corresponding LiDAR (or depth camera) measurement. 
We only store the ORB feature points that have the corresponding 3D coordinates.

\textit{Compact image descriptor VLAD.} Based on the 64 representative feature descriptors in the visual vocabulary, we calculate a VLAD matrix of each reference image as follows.\\
\textbf{(i) [Feature extraction]} Using the ORB feature detector, We extract multiple feature descriptors from the image.  \\
\textbf{(ii) [Cluster extracted ORB feature descriptors]} Using the trained k-means classifier described in \textit{B. Step 1: Visual vocabulary construction}, we assign all ORB feature descriptors to one of the 64 clusters. \\
\textbf{(iii) [Calculate errors for each cluster]} We calculate an error between every ORB feature descriptor in the k-th cluster and the center of the k-th cluster, which is one of 64 representative feature descriptors in the visual vocabulary.
We call this error as a residual error. \\
\textbf{(iv) [Unnormalized VLAD matrix]} We sum all residual errors allocated to the k-th cluster, i.e., we can get 64 vectors where each vector is a sum of all residual errors in the k-th cluster. 
When we stack these vectors, we can get a 64-by-32 matrix for each image (64 is the number of the clusters, and 32 is the dimension of the ORB descriptor). \\
\textbf{(v) [Normalized VLAD matrix]} The calculated 64-by-32 matrix is then L2-normalized row-wise (intra-normalization) and finally L2-normalized in its entirety. \\
The output from this process is called as a VLAD matrix. More details about VLAD can be found in \cite{arandjelovic2013all} and \cite{jegou2010aggregating}.

\textit{Global pose of the image.} It should be pointed out that the mapping process in this paper needs independent positioning device or algorithm to obtain the global pose of the image. For example, in the experiments presented in this work, we use differential GPS or indoor positioning system for mapping to get the global pose of each reference image. 

After we obtain the information of each reference image, we aggregate the information and build the map. 
For each image, we can obtain the information as a tuple, (extracted ORB feature descriptors with 3D coordinates, VLAD matrix, global pose).
By simply stacking them, we can get a set of these tuples \{(extracted ORB feature descriptors with 3D coordinates, VLAD matrix, global pose)\} which we will refer to  as the ``map database".

In the localization step, we search tuples from the map database which are similar to a tuple that contains a query image's information. 
To search relevant tuples from the map, we train the k-nearest neighbor (KNN) classifier on the VLAD matrices of the map so that we can find similar tuples in terms of VLAD matrix when the query VLAD matrix is given.
k is set to 10 for KNN in our experiments.

\subsection{Step3: Localization}

\begin{figure}[ht]
\begin{center}
\includegraphics[width=1\linewidth,keepaspectratio]{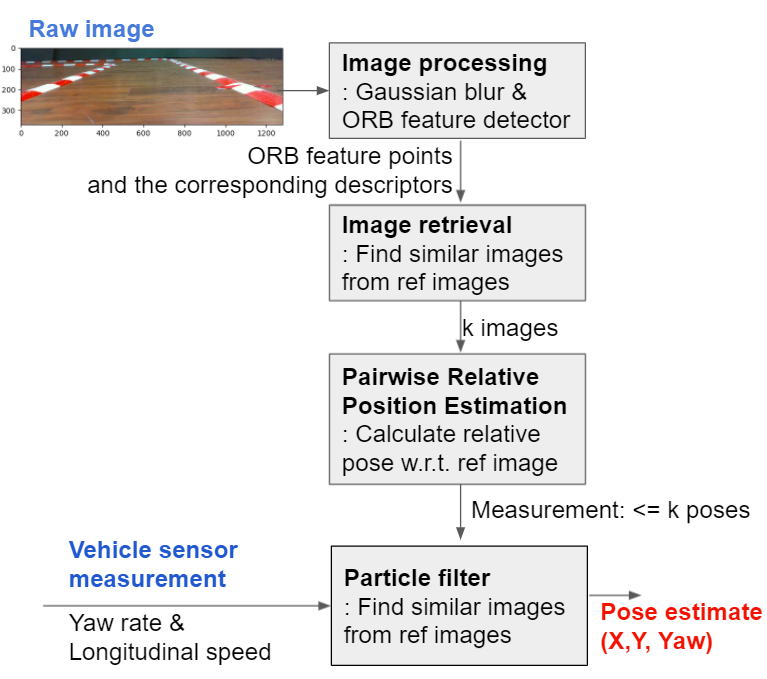}
\caption{Overall localization architecture}
\label{architecture}
\end{center}
\vspace{-1.0em}
\end{figure}

The overall localization architecture is given in Fig. \ref{architecture}.
The localization in our method consists of three consecutive steps.
First, the localization module finds k reference images from the map which are similar to a query image using the image retrieval method. 
Subsequently, the module calculates the relative pose between the query image and each retrieved reference image.
Since we have the corresponding global pose for each retrieved reference image, we obtain k hypothesis about the query image's global pose as an output of the second sequence.
Finally, using k hypothesis and on-board vehicle sensor signals as inputs, we design a particle filter to estimate the global pose of the query image and thus the global pose of the vehicle.
While the mapping process needs multiple sensors, the localization step can be done with a single camera. 
In the rest of this section, details of each sequence will be described. 

\textbf{Image processing.}
To suppress the effect of the noise, we apply a 5\(\times\)5 Gaussian blur filter to raw image.
Then, the algorithm extracts ORB feature descriptors and compresses a query image into a VLAD matrix

\textbf{Image Retrieval to find 10 Reference Images Similar to a Query Image.}
By using the KNN classifier described in the previous section \textit{C. Step2: Mapping}, we can find 10 tuples from the map which are similar with the query image's tuple in terms of VLAD matrix.

\textbf{Pairwise Relative Pose Estimation to calculate the Relative Pose between the Query and each Retrieved Reference Image.}
In the previous step, we found the 10 similar tuples from the map which contains extracted ORB feature descriptors with 3D coordinates. 
With these feature descriptors and their 3D position, the relative pose between query and reference images can be estimated by feature matching and solving a Perspective-n-Point problem (PnP) \cite{li2012robust}.

Feature matching is the process to find 2D point correspondence between query image's feature descriptors and the reference image's feature descriptors.
In this paper, 2D point correspondence are found with a brute-force matcher \cite{rublee2011orb}.

After the 2D point correspondence is solved, we solve the PnP problem given the  3D coordinates of the matched feature descriptors. 
By solving a PnP problem with the Random sample consensus (RANSAC) scheme, the relative pose between the query image and each retrieved reference image can be obtained.
Sometimes, the image retrieval block retrieves the reference image that is not similar with the query image because we compress an image into a low dimensional matrix.
In this case, PnP RANSAC logic fails to estimate the global pose.
Failure of PnP RANSAC here means that there is no inlier or the calculated relative angle is larger than camera's field of view.
In that case, we remove the tuple of such reference image from the set of k tuples.

Since we have the corresponding global pose for each retrieved reference image, we obtain \(\leq\)k hypothesis about the query image's global pose after matching feature descriptors and solving a PnP problem.

\textbf{Particle Filter to  estimate the global pose of the vehicle.}
Using multiple hypothesis and on-board vehicle sensor signals as inputs, we design a particle filter to estimate the global pose of the query image, and thus the global pose of the vehicle.
A particle filter is the Bayes filter that represents the posterior probability distribution with a set of samples drawn from the distribution \cite{thrun2002probabilistic}.
These samples of a posterior distribution are called particles, and each particle denotes a hypothesis as to what the true state may be.
Since a particle filter is the Bayes filter, we need to design a measurement model and a motion model to construct the particle filter.
In the rest of the section, we describes the details of designing the measurement model and the motion model.

The measurement model is a probability density function \(p(z|x)\) where \(z\) is a measurement and \(x\) is a state of the query image.
In a particle filter algorithm, each particle represents the estimate of the state, and \(x\) is a state of the particle.
Thus, \(p(z|x)\) is the probability of the measurement \(z\) under the particle's state \(x\).
The measurement \(z\) is calculated by averaging all global pose estimates of the query image from the previous module, Pairwise Relative Pose Estimation, with the RANSAC scheme.
The particle's state \(x\) consists of the VLAD and the global pose of the query image.

To design the measurement model of the particle filter, we use a Gaussian Mixture Model (GMM).
GMM are linear combinations of Gaussian distributions.
Thus, we need to design each Gaussian distribution and the corresponding coefficient.

We design the Gaussian distribution based on the output of the Pairwise Relative Pose Estimation block.
Pairwise Relative Pose Estimation provides multiple (equal to or less than k) hypothesis on the global pose of a query image. 
Since each hypothesis is an estimate of the query image's global pose, it is likely that the actual global pose of the query image is in the vicinity of each hypothesis.
To model this probability, we use Gaussian distribution as follows.
\begin{equation}\label{gaussian component}
\begin{split}
& p(z|\text{i-th retrieved image}) \sim N(z_{\text{i-th retrieved}} + \mu_{1}, \Sigma_{1}) \\
& z=\left[ \begin{matrix}
   X,
   Y,  
   \psi
\end{matrix} \right]^{T} \\
\end{split}
\end{equation} 
where \(X\) and \(Y\) are X, Y coordinates in global frame respectively, \(\psi\) is the heading angle, and \(z_{\text{i-th retrieved}}\) is an estimate of query image's global pose based on i-th retrieved image's information.
Since there are multiple hypothesis of query's global pose, which is calculated from the pairwise relative pose estimation block, we have multiple Gaussian distributions that correspond to each hypothesis.

The mean vector \(\mu_{1}\) and the covariance matrix \(\Sigma_{1}\) of the Gaussian distribution model in (\ref{gaussian component}) are calculated with separate training dataset. 
When a query image is drawn from this training dataset, we directly know k-closest images in terms of distance. 
Matching feature descriptors and solving a PnP problem, we can get k estimates of the global pose of the query image. 
By using these estimates, we can obtain Maximum Likelihood Estimate (MLE) of the mean vector \(\mu_{1}\) and the covariance matrix \(\Sigma_{1}\), which are:
\begin{equation}\label{MLE}
\begin{split}
& e_{i} = z_{\text{query}} - z_{\text{i-th retrieved}} \\
& \hat{\mu_{1}} = \frac{\sum\nolimits_{i=1}^{k}{{{e}_{i}}}}{k}, \hat{\Sigma_{1}} = \frac{\sum\nolimits_{i=1}^{k}{{{e}_{i}}{{e}_{i}}^{T}}}{k} \\
\end{split}
\end{equation}
where \(z\) is a vector defined in (\ref{gaussian component}).
Note that we know the global pose of the query image, \(z_{\text{query}}\), exactly in this case because this query image is drawn from the training set where all information including global pose are accessible.

We now have multiple Gaussian distributions which are in the form of (\ref{gaussian component}) and with MLE parameter estimates in (\ref{MLE}) as parameters. 
Each distribution represents the probability density function whose value represents the likelihood that the measurement \(z\) is the true global pose of the query image given the retrieved image.
The remaining problem is how to integrate these multiple distributions in one probability density function which will be the measurement model of the particle filter.
Since we use a GMM to design the measurement model, this problem can be casted as the  computation of a coefficient which  corresponds to each Gaussian distribution.

We design the corresponding coefficient as 
\begin{equation}\label{GMM_coeff}
c_{i} = p(\text{i-th retrieved image}|\text{Query image}, \text{Pose};\text{Map})
\end{equation}
where Query image denotes the relevant information that can be obtained from the query image, and is a VLAD of the query image in this paper; Pose denotes the global pose \(\left[ \begin{matrix}
   X,
   Y,  
   \psi
\end{matrix} \right]\) of the query image; and Map denotes the set of tuples previously described in the section \textit{C. (Step2: Mapping)}.
This coefficient is the conditional probability of retrieving the specific reference image from the map given the query image and the pose of the query image.

There are two main variables which affect this conditional probability of retrieving the correct reference image: the VLAD matrix and the the pose.
The similarity in terms of VLAD matrix affects the probability in~(\ref{GMM_coeff}) since this is used by the how the image retrieval method to finds similar images from the map. 
As the disparity of VLAD matrix increases, the probability of the retrieving the certain image decreases.
Second, the pose closeness affects the probability in~(\ref{GMM_coeff}) .
For example, when the pose \(\left[ \begin{matrix} X,Y,\psi\end{matrix} \right]\) of the certain image is close enough to that of the query image, it is highly likely two images are similar with each other.
In this sense, the coefficient of GMM, \(c_{i}\) is in (\ref{GMM_coeff}), is modelled in the forms of Gaussian distribution as
\begin{equation}\label{coefficient component}
\begin{split}
& p(\text{i-th retrieved image}|\text{Query image}, \text{Pose};\text{Map}) \\
& = det(2 \pi \Sigma_{2})^{-\frac{1}{2}} \, exp(-\frac{1}{2}(e-\mu_{2})^{T}\Sigma_{2}^{-1} (e-\mu_{2})) \\
& e=\left[ \begin{matrix}
   e_{vlad} \\
   X_{\text{i-th retrieved}} - X \\
   Y_{\text{i-th retrieved}} - Y \\
   \psi_{\text{i-th retrieved}} - \psi \\
\end{matrix} \right] \\
& e_{vlad} = \left\| {{vlad}_\text{i-th retrieved}} - {{vlad}_{\text{query}}} \right\|_{F}  \\
\end{split}
\end{equation}
where \(vlad_{*}\) is the VLAD matrix of * image, and \(\left\| * \right\|_{F}\) is Frobenius norm.
The mean vector \(\mu_{2}\) and the covariance matrix \(\Sigma_{2}\) of the Gaussian distribution model are calculated in same way as described in (\ref{MLE}) by augmenting one additional variable \(e_{vlad}\).

Finally, the GMM can be defined as:
\begin{equation}\label{GMM}
\begin{split}
& p(z|\text{Query image}, \text{Pose};\text{Map}) \\
& = \sum\limits_{all\ i}{c_{i} \cdot p(z|\text{i-th retrieved image})}  \\
\end{split}
\end{equation}
where \(c_{i}\) is in (\ref{coefficient component}). 
Suppose both the VLAD and the global pose of the query image is a state vector, i.e., a set of data describing exactly where an object is located in space.
Then, the conditional probability in (\ref{GMM}) can be written as \(p(z|x)\) in (\ref{GMM_MeasurementModel}), which is a measurement model.
\begin{equation}\label{GMM_MeasurementModel}
\begin{split}
& p(z|x) = p(z|\underbrace{\text{Query image}, \text{Pose}}_{x};\text{Map}) \\
& x=\left[ \begin{matrix}
   vlad_{\text{query}},
   X,
   Y,
   \psi
\end{matrix} \right]^{T} \\
\end{split}
\end{equation}
In the particle filter, \(x\) is a state of each particle, and \(vlad_{\text{query}}\) is set to be identical for all particles.
In contrast, the pose \(\left[ \begin{matrix} X,Y,\psi\end{matrix} \right]\) of each particle is generally different and is evolved based on the governing probabilistic equation of the  motion model. 

For the motion model of the particle filter, we use a vehicle kinematic model with longitudinal speed and yaw rate as its inputs. The kinematic model utilized in this paper is:
\begin{equation} \label{motion model}
\begin{split}
 & \dot{X}(t)=\hat{V}(t)\cos \psi(t)  \\ 
 & \dot{Y}(t)=\hat{V}\cos \psi(t)  \\ 
 & \dot{\psi }(t)=\hat{\gamma }(t) \\ 
 & \hat{V}(t)={{V}_{meas}(t)}+{{w}_{v}(t)},\ \ {{w}_{v}(t)} \sim U[\underline{v},\bar{v}] \\ 
 & \hat{\gamma }(t)={{\gamma }_{meas}(t)}+{{w}_{\gamma }(t)},\ \ {{w}_{\gamma }(t)}\sim U[\underline{\gamma },\bar{\gamma }] \\ 
\end{split}
\end{equation}
where \(X\) and \(Y\) are X, Y coordinates of the vehicle in global frame respectively, \(\psi\) is the heading angle of the vehicle, \({V}_{meas}\) is measured longitudinal speed, \({\gamma }_{meas}\) is measured yaw rate, and \({w}_{v}\) and \({w}_{\gamma }\) are additional random terms to represent measurement noise. The noise terms are designed as uniform distribution.
To utilize this kinematic model in particle filter, we discretize the model in the Euler method.
Note that any model with bounded error can be used as a motion model.

In summary, by using the kinematic model in (\ref{motion model}) as the motion model and the GMM in (\ref{GMM}) and (\ref{GMM_MeasurementModel}) as the measurement model, we devise a particle filter to estimate the global pose.

\section{Experiments}
We investigate the proposed localization method using KITTI dataset \cite{Geiger2012CVPR}. Then, we conduct indoor closed-loop tests with an 1:10 scale test platform. 
For both experiments, the localization logic is executed on a host laptop with an Intel(R) Core(TM) i7-9750H CPU and 32GB memorywithout support of GPU hardware.
Given this configuration, our method runs at approximately 10Hz.
The worst case runs at 5Hz.

\subsection{KITTI odometry dataset}
\textbf{Test setting:} The dataset we used is KITTI odometry dataset \cite{Geiger2012CVPR}.
All sequences in this dataset have images and LiDAR point clouds.
We employ two sequences among 10 sequences of the KITTI odometry dataset with the global pose. 

\begin{figure}[ht]
\begin{center}
\includegraphics[width=1\linewidth,keepaspectratio]{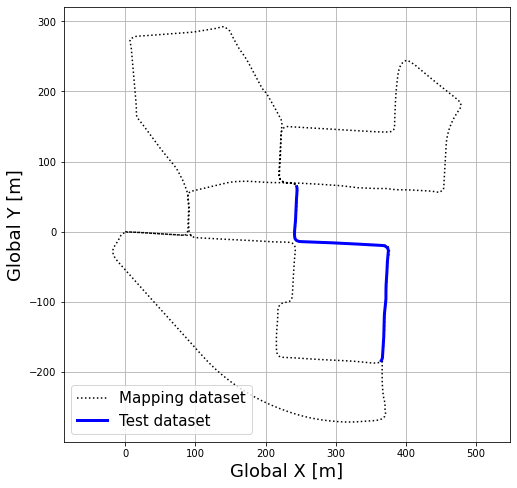}
\caption{KITTI odometry dataset: Sequence 00}
\label{KITTI}
\end{center}
\vspace{-1.0em}
\end{figure}

We use one sequence for the training stage(to train visual vocabularies and parameters in GMM model), and split another sequence into two parts: \(\sim\)85 \% for the mapping stage and \(\sim\)15 \% for the test stage as illustrated in Fig. \ref{KITTI}. The test road section which is denoted by the blue lines in Fig. \ref{KITTI} is the road section where the vehicle passed the same region twice.

For mapping, we utilize all measured data from the mapping dataset: global pose from differential GPS, LiDAR point clouds, and images. For localization, we only use images sequentially drawn from the testing dataset. 

\begin{figure}[ht]
\vspace{-0.5em}
\begin{center}
\includegraphics[width=1\linewidth, height=1\linewidth,keepaspectratio]{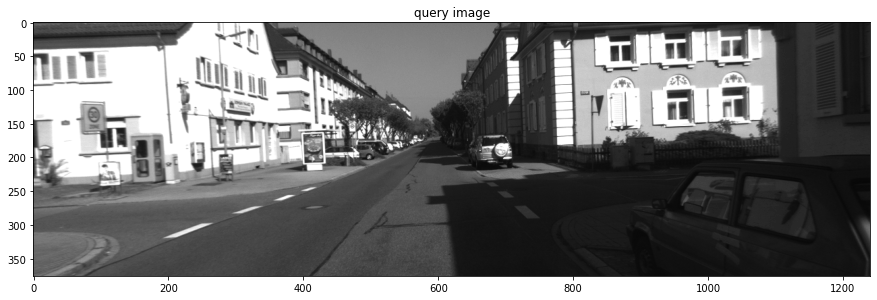}
\caption{Query Image}
\label{query}
\end{center}
\vspace{-1.0em}
\end{figure}

\begin{figure}[ht]
\vspace{-0.5em}
\begin{center}
\includegraphics[width=1\linewidth,keepaspectratio]{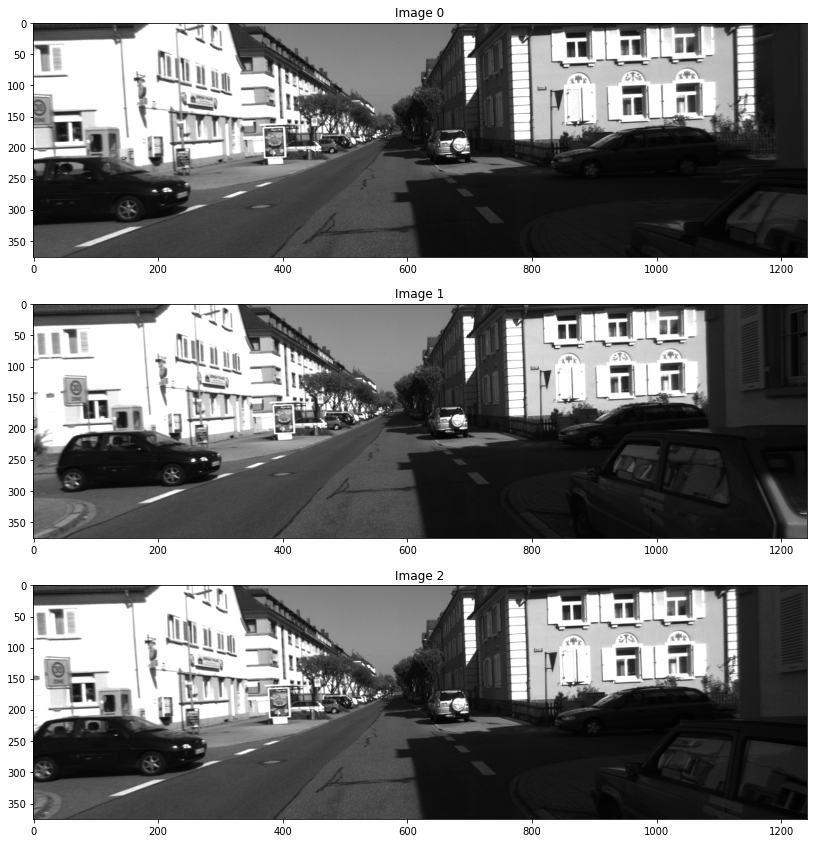}
\caption{Image retrieval: top 3 matches}
\label{top5}
\end{center}
\vspace{-1.0em}
\end{figure}

\textbf{Result and Discussion - Image retrieval.}
We present the results of the image retrieval method and discussion on the results.
Fig. \ref{query} is the query image. 
Fig. \ref{top5} show top 3 matching reference images which show that these retrieved images are correctly localized.
Note that the reference images are just for visualization and not a part of the map.

As illustrated in Fig. \ref{top5}, even if there are two cars which do not exist in the query image, the image retrieval works well.
We test the image retrieval logic for all cases, and throughout the cases at least 3 images among 5 matches are well retrieved. 

There are two reasons why this simple image retrieval logic works well for localization of automated vehicles.
First, the viewpoint of the camera is consistent when vehicles are driving on the road.
It is reported that the image retrieval performance can be improved when the viewpoint of the reference images are similar with that of the query image \cite{torii201524}.
Since it is difficult to take a picture in the similar viewpoint with hand-held device, \cite{torii201524} proposes the synthetic view method that transform the viewpoint of the image.
However, in localization of automated vehicles, this image viewpoint transform is less effective and might be avoided.
In fact, because automated vehicles driving at moderate speed on public roads always align with the road direction, the images taken from the automated vehicles have similar viewpoint with the images taken from other vehicles for mapping.

Second, we note that the illumination conditions in the dataset are sufficient for the image retrieval.
KITTI dataset only contains the images with good, bright illumination condition.
However, multiple researches \cite{torii201524, arandjelovic2016netvlad, torii2019large} point out that dark illumination condition deteriorates the image retrieval performance.
This issue could be alleviated if images in the night time are collected and added to the map as in \cite{burki2019vizard}.
In fact, the map is in the form of the database so new images can be easily added to the map database even if new images have the same poses as the existing reference images'.
In other words, unlike 3D map based approach, the map database approach in this paper can register multiple images on the same global location.
Using the map database, the image retrieval method can search reference images from the database that is similar with the query image, i.e., if the query image's illumination condition is dark, then the image retrieval is able to find similar, dark reference images from the map database.

\textbf{Result and Discussion - Localization:}
The overall performance of the localization module is presented in this section.

\begin{figure}[ht]
\vspace{-0.5em}
\begin{center}
\includegraphics[width=1\linewidth, height=1\linewidth,keepaspectratio]{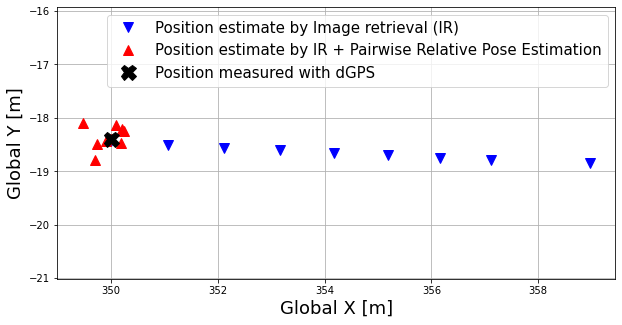}
\caption{Result: Pairwise Relative Pose Estimation}
\label{relpe}
\end{center}
\vspace{-0.5em}
\end{figure}

Fig. \ref{relpe} shows the reason why pairwise relative position estimation is needed. 
The image retrieval method only can give an approximate camera pose estimate as illustrated in blue points of Fig. \ref{relpe}. 
Simply averaging these blue points is not sufficient because the average point can be biased(ex: In Fig. \ref{relpe}, the points located in the right side of the ground truth outnumber the points in the left side) and outliers will degrade the performance.
By conducting pairwise relative pose estimation, we can get more accurate version of camera pose estimate as illustrated in red points of Fig. \ref{relpe}.

\begin{figure}[ht]
\vspace{-0.5em}
\begin{center}
\includegraphics[width=1.2\linewidth, height=1.2\linewidth,keepaspectratio]{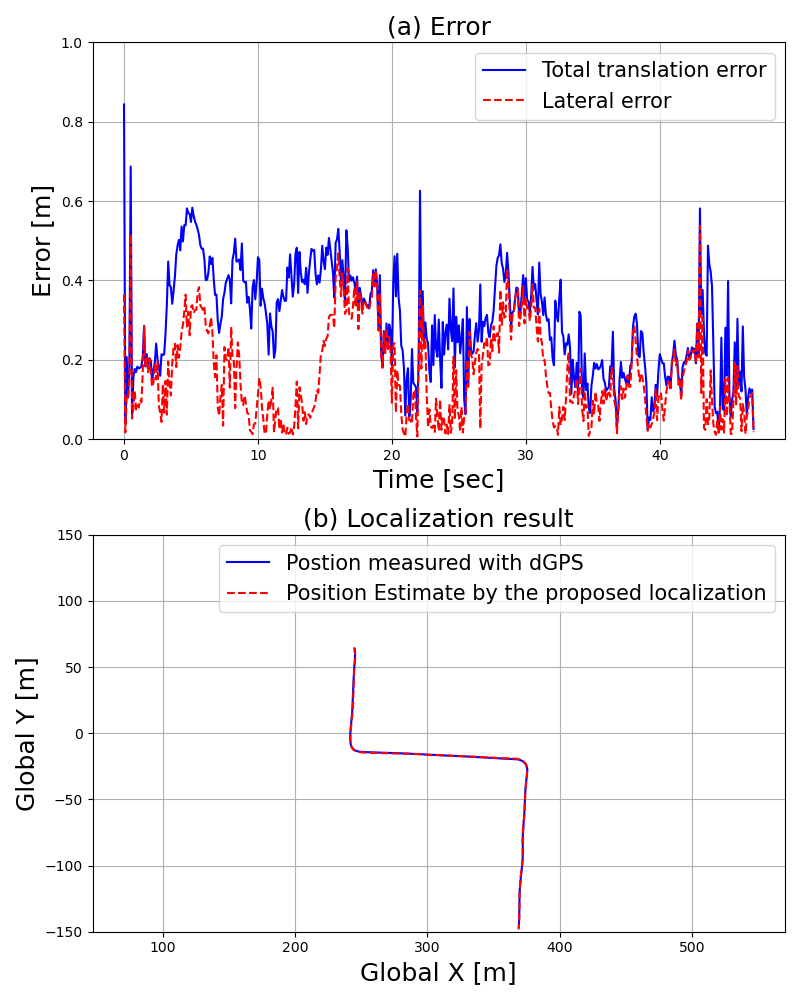}
\caption{Result: Overall localization performance}
\label{traj}
\end{center}
\vspace{-0.5em}
\end{figure}
Fig. 6 shows the overall localization performance. 
The Root Mean Square Error(RMSE) of the performance is 0.313m, which is comparable to the performance of monocular camera localization with given 3D LiDAR map in \cite{caselitz2016monocular} (average translational error for KITTI odometry dataset 00 in \cite{caselitz2016monocular}:  0.30m\(\pm\)0.11m, reported maximum error 0.8m, minimum error 0.2m).

\subsection{Indoor testing with 1:10 Autonomous Vehicle Platform}
\textbf{Test setting:} We investigate our method on 1:10 Berkeley Autonomous Race Car (BARC) platform in a laboratory race track environment. 
The car is equipped with an Intel RealSense d435i camera and a HTC Vive Tracker for indoor positioning system.
The localization logic is executed run on a host laptop with an Intel(R) Core(TM) i7-9750H CPU and 32GB memory without
support of GPU hardware, and the host laptop send the global pose of the vehicle via Wifi network.
For mapping, we use global pose of the vehicle from Vive tracker and both depth and color images from RealSense d435i.
During mapping, the vehicle tracks a center line of the indoor racetrack for multiple laps and records all poses, images, and depth images to construct the map.
For localization, we use the constructed map, yaw rate, longitudinal speed, and color image from RealSense d435i.

We investigate our localization logic in two ways. 
First, we test open-loop case. 
We calculate the vehicle's pose with the proposed localization logic but we do not feed the estimate back to the vehicle.
In this open-loop test case, we feed vehicle's pose from the Vive tracker back to the vehicle for tracking the reference path.
Second, we test closed-loop case.
In this closed-loop test case, we calculate vehicle's pose with the proposed localization logic and use the estimate for tracking the reference path.
For both cases, we use the same Proportional-Integral-Derivative (PID) trajectory tracking controller with longitudinal speed, lateral and heading errors as an error.
The reference path is a center line of the racing track, and the reference speed is set to be constant.

\begin{figure}[h]
\vspace{-0.5em}
\begin{center}
\includegraphics[width=1.0\linewidth, height=1.0\linewidth,keepaspectratio]{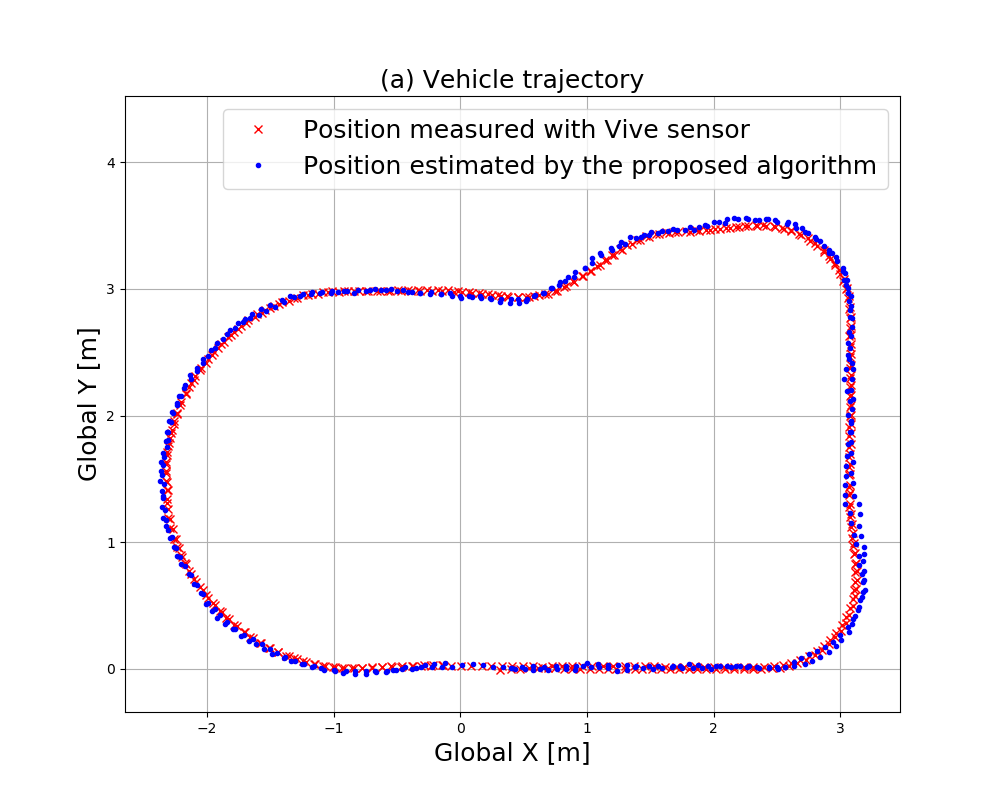}
\caption{Open-loop: Comparison the proposed localization with Vive tracker}
\label{openloop}
\end{center}
\vspace{-0.5em}
\end{figure}

\begin{figure}[h]
\vspace{-0.5em}
\begin{center}
\includegraphics[width=1\linewidth, height=1\linewidth,keepaspectratio]{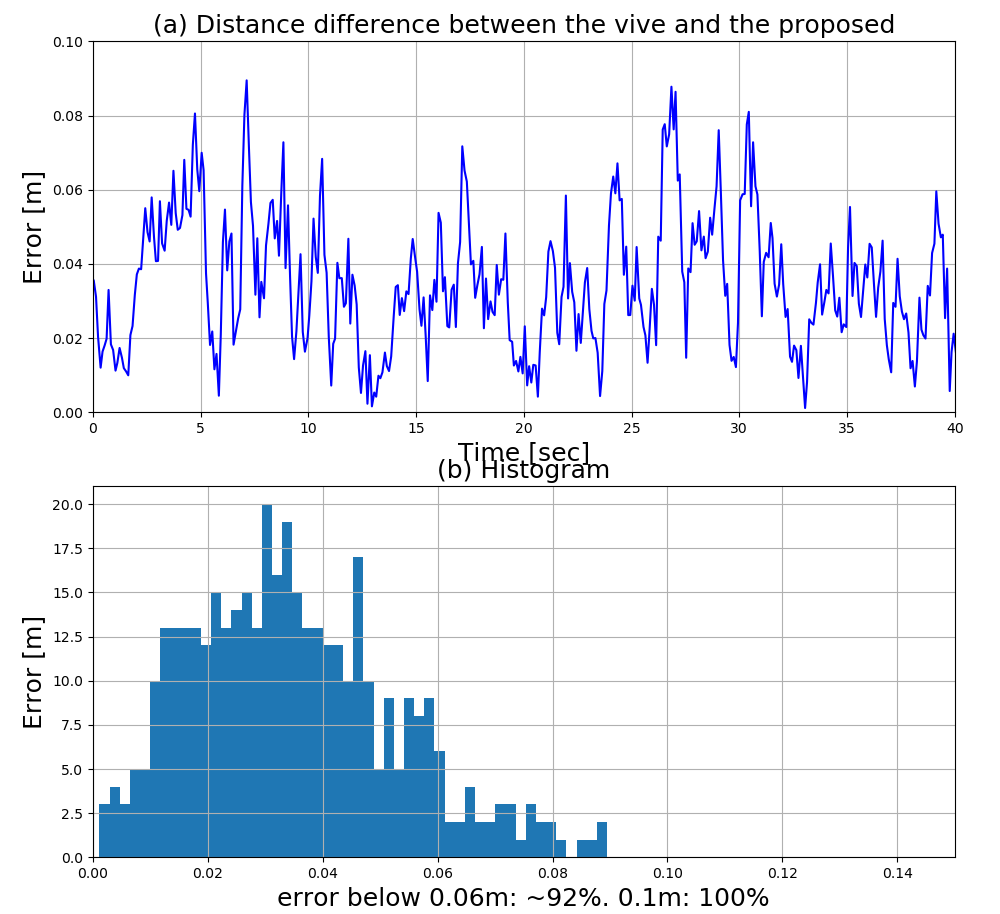}
\caption{Open-loop: Localization error analysis}
\label{openloop_distance}
\end{center}
\vspace{-1.5em}
\end{figure}

\textbf{Result and Discussion - Open loop.}
Fig.\ref{openloop} and Fig.\ref{openloop_distance} present the results of open-loop test.
As illustrated in Fig.\ref{openloop}, the recorded trajectories overlap.
The estimation errors can be calculated with the pose measurements from the Vive tracker as the ground truth.
For the open-loop case, estimation errors are below 0.1m as illustrated in Fig.\ref{openloop_distance}. (100 \% below 0.1m, \(\sim\)92 \% below 0.06m).

\begin{figure}[h]
\vspace{-0.5em}
\begin{center}
\includegraphics[width=1\linewidth, height=1\linewidth,keepaspectratio]{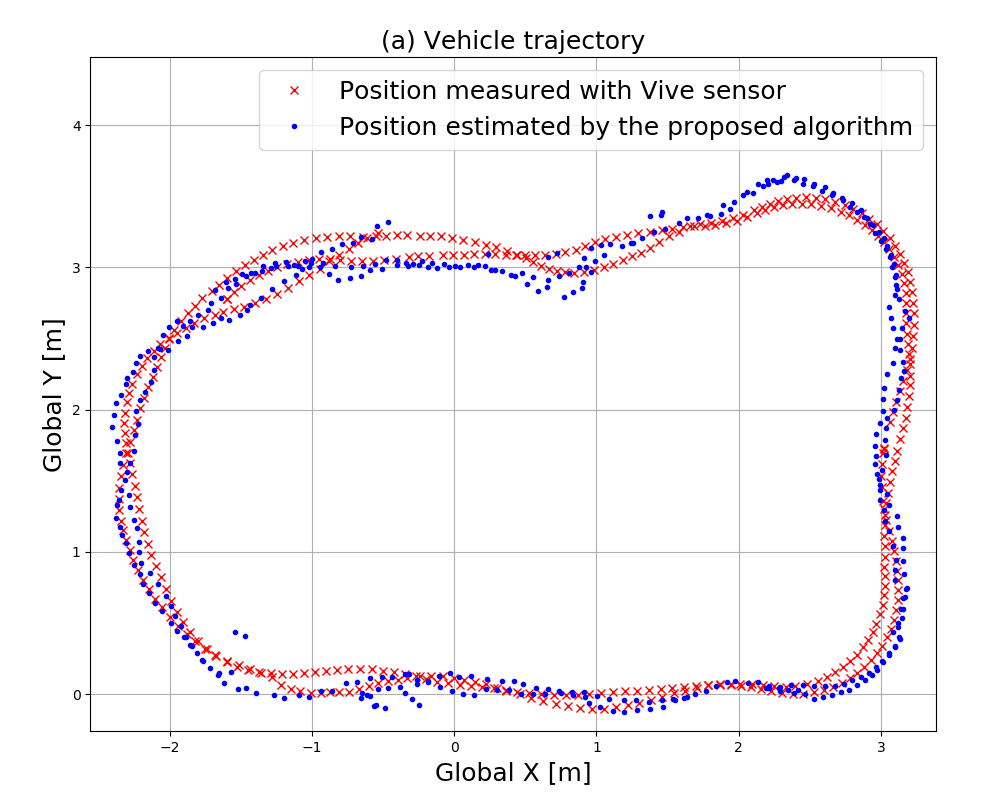}
\caption{Closed-loop: Comparison the proposed localization with Vive tracker}
\label{closedloop}
\end{center}
\vspace{-1.0em}
\end{figure}

\begin{figure}[h]
\vspace{-0.5em}
\begin{center}
\includegraphics[width=1\linewidth, height=1\linewidth,keepaspectratio]{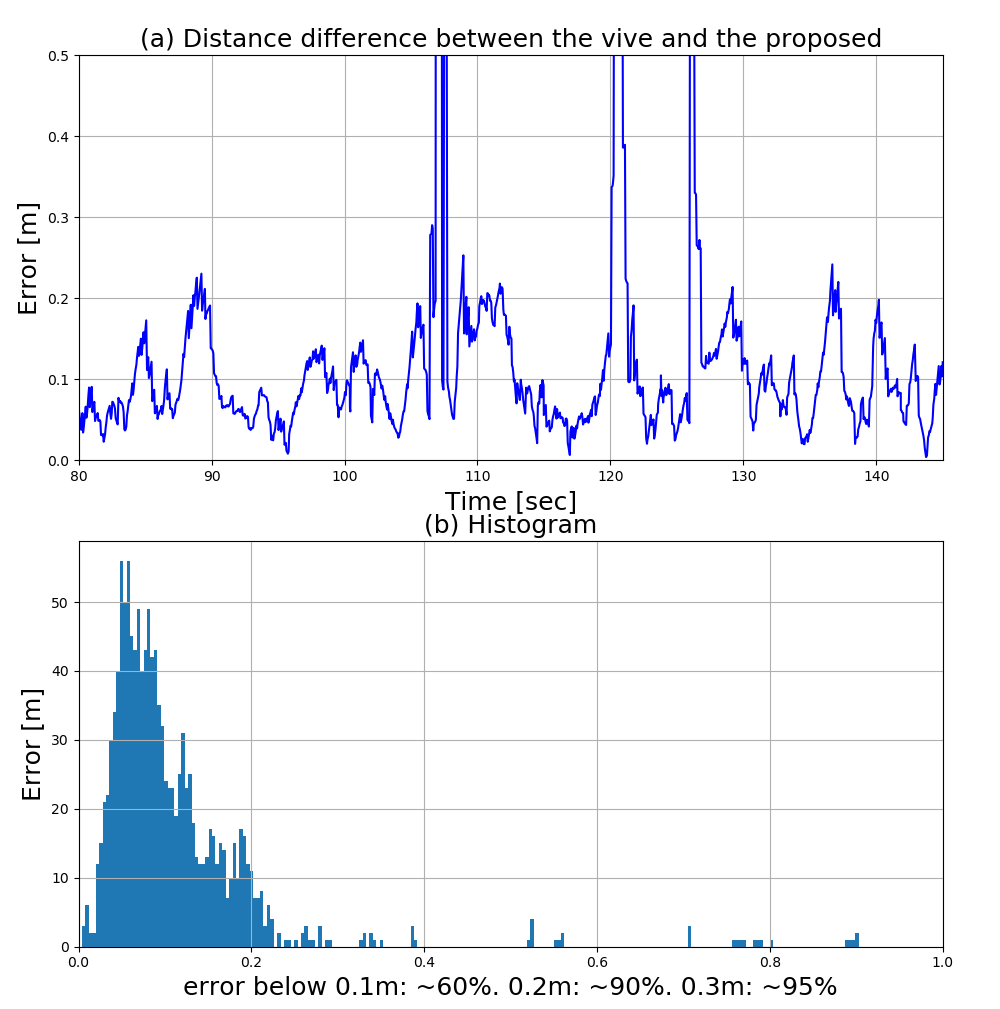}
\caption{Closed-loop: Localization error analysis}
\label{closedloop_distance}
\end{center}
\vspace{-1.5em}
\end{figure}

\textbf{Result and Discussion - Closed loop.
} Fig.\ref{closedloop} and Fig.\ref{closedloop_distance} present the results of closed-loop test.
There are few intervals where the distance error increases, i.e., high error of X:  2.5m\(\sim\)3.5m, Y: 0m\(\sim\)1m in Fig.\ref{closedloop}.
This is the interval where the vehicle encountered a featureless wall.
In this case, the localization logic relies only on propagating motion model, i.e., Dead-reckoning.

As illustrated in Fig.\ref{closedloop_distance}, in contrast to the results of the open-loop test, the localization error increases. 
We conjecture that this is due to a positive feedback loop between localization and control errors.
As localization error increases, tracking error increases.
As the control error increases, the camera mounted on the vehicle can face the environment that was not mapped. 
In other words, the image contains some features that were not a part of the map.
These new features in the image mean that the ratio of the known features, which are part of the map, decreases. 
Since the localization logic, especially the pairwise relative pose estimation, relies on the recorded features in the map, the decreased ratio of the known features deteriorates the localization error.
As a result of the positive feedback loop, the localization performance in Fig.\ref{closedloop} is worse than that in Fig.\ref{openloop}.

To experimentally prove this conjecture, we further investigate our algorithm through additional closed-loop tests by augmenting the map database. 
If our conjecture is true, then we can alleviate the positive feedback loop introduced by the localization algorihtm by adding  data corresponding to  large heading error between the center line and vehicle orientation  to the map database after the first few laps. 
Thus, on top of the existing map database, we add data which was recorded when the vehicle is deliberately controlled in an oscillatory motion to generate large heading errors.
To deliberately control the vehicle to show large heading errors, we add a sinusoidal disturbance to a steering control input of the existing PID controller.

\begin{figure}[h]
\vspace{-0.5em}
\begin{center}
\includegraphics[width=1\linewidth, height=1\linewidth,keepaspectratio]{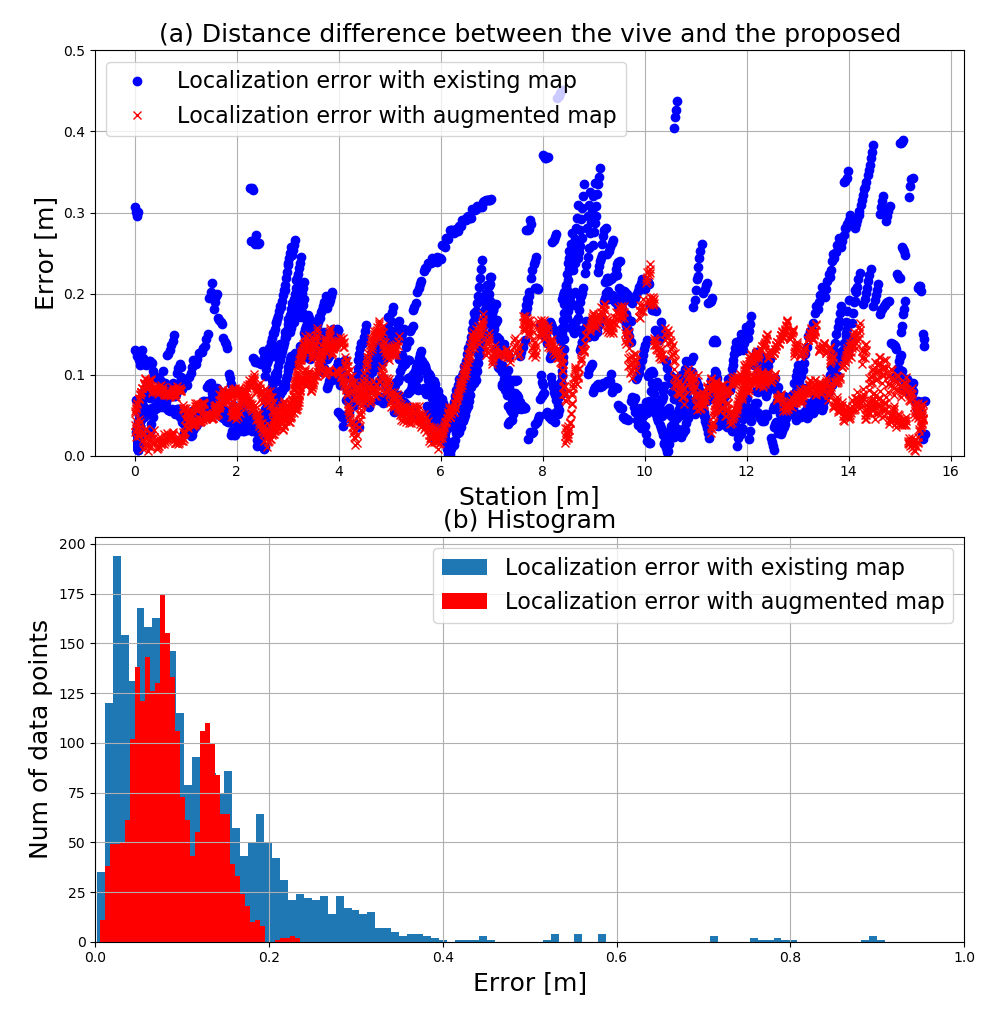}
\caption{Closed-loop tests: Experimentally prove the conjecture of the positive feedback loop between localization errors and control errors}
\label{aug_difference}
\end{center}
\vspace{-1.5em}
\end{figure}

Fig. \ref{aug_difference} shows the results of the closed-loop tests with the augmented map. 
The localization algorithm with the augmented  map database shows better position estimation performance compared to the same algorithm using the initial map database.
Remarkably, not only does the overall error decreases, but also the spikes in the localization error of the initial tests as 107sec, 121sec, and 127sec in Fig.\ref{closedloop_distance} disappear.

\section{Conclusion}
In this paper, we have presented a new monocular camera localization method for autonomous vehicles using image retrieval method.
The proposed method consists of three stages: (1) Visual vocabulary construction, (2) Mapping, (3) Localization.
While mapping process needs multiple sensors, localization can be done with a single camera.
We investigated the proposed localization method using KITTI dataset and via closed-loop indoor test with 1:10 Autonomous vehicle platform. 
The results has shown the effectiveness of the proposed localization method and real-time operating capability of the visual localization module.
Also,  from  the  experiment  results  of closed-loop indoor tests, we show a new problem of a positive feedback loop between a localization error and a control error.
Since the proposed monocular camera localization method has shown good localization performance in open-loop test, we aim to develop robust model predictive control to alleviate a positive feedback between localization and control errors.

\bibliographystyle{ieee_fullname}
\bibliography{egbib}

\end{document}